\pdfoutput=1
\documentclass[11pt,a4paper]{article}
\usepackage[hyperref]{naaclhlt2019}
\usepackage{times}
\usepackage{latexsym}
\usepackage{url}
\usepackage{amssymb}
\usepackage{amsmath}
\usepackage{graphicx}
\usepackage{booktabs}
\usepackage{mathtools}
\usepackage{fancyhdr}
\usepackage{xcolor}
\usepackage{natbib}
\bibpunct{(}{)}{;}{a}{,}{,}
 \relax
\setcitestyle{notesep={: },yysep={, }} \relax
\usepackage[utf8]{inputenc}
\usepackage{amsmath}
\usepackage{stmaryrd}
\usepackage{pst-node}
\usepackage{tikz-cd} 
\usepackage{multicol}
\usepackage{sidecap}
\usepackage{subcaption}
\usepackage{listings}

\aclfinalcopy 
\def\aclpaperid{782} 

\newcommand{\Speaker}{S}
\newcommand{\Listener}{L}

\newcommand{\SZEROWORD}{\Speaker_{0}^{\text{WD}}}
\newcommand{\SONEWORD}{\Speaker_{1}^{\text{WD}}}
\newcommand{\SONEWORDC}{\Speaker_{1}^{\text{WD-C}}}

\newcommand{\LZEROWORD}{\Listener_{\text{0}}^{\text{WD}}}
\newcommand{\LONEWORD}{\Listener_{\text{1}}^{\text{WD}}}

\newcommand{\SZEROSENT}{\Speaker_{\text{0}}^{\text{SNT}}}
\newcommand{\SONESENTGP}{\Speaker_{\text{1}}^{\text{SNT-GP}}}
\newcommand{\SONESENTIP}{\Speaker_{\text{1}}^{\text{SNT-IP}}}

\newcommand{\SONESENTCGP}{\Speaker_{\text{1}}^{\text{SNT-CGP}}}
\newcommand{\SONESENTCIP}{\Speaker_{\text{1}}^{\text{SNT-CIP}}}

\newcommand{\LZEROSENT}{\Listener_{\text{0}}^{\text{SNT}}}
\newcommand{\LONESENT}{\Listener_{\text{1}}^{\text{SNT}}}

\usepackage{color}

\title{Lost in Machine Translation: A Method to Reduce Meaning Loss} 
\date{}
\author{Reuben Cohn-Gordon \\
  Stanford 
  \\\And
  Noah D. Goodman \\
  Stanford \\
  }

\begin{document}
\maketitle
\begin{abstract}

A desideratum of high-quality translation systems is that they preserve meaning, in the sense that two sentences with different meanings should not translate to one and the same sentence in another  language. However, state-of-the-art systems often fail in this regard, particularly in cases where the  source and target languages partition the “meaning space” in different ways. For instance, ``I cut my finger.'' and ``I cut my finger off.'' describe different states of the world but are translated to French (by both \emph{Fairseq} and \emph{Google Translate}) as ``Je me suis coup\'e le doigt.'', which is ambiguous as to whether the finger  is  detached. More generally, translation systems are typically many-to-one (non-injective) functions from source to target language, which in many cases results in important distinctions in meaning being lost in translation. Building on Bayesian models of informative utterance production, we present a method to define a less ambiguous translation system in terms of an underlying pre-trained neural sequence-to-sequence model. This method increases injectivity, resulting in greater preservation of meaning as measured by improvement in cycle-consistency, without impeding translation quality (measured by BLEU score).

\end{abstract}

\section{Many-to-One Translations}\label{brown}



Languages differ in what meaning distinctions they must mark explicitly. As such, translations risk mapping from a form in one language to a more ambiguous form in another. For example, the definite (\ref{sent:3}) and indefinite (\ref{sent:4}) both translate (under \emph{Fairseq} and \emph{Google Translate}) to (\ref{sent:5}) in French, which is ambiguous in definiteness.

    


    


\begin{align}
    \emph{The animals run fast.} \label{sent:3} \\ 
    \emph{Animals run fast.} \label{sent:4} \\
    \emph{Les animaux courent vite} \label{sent:5}
\end{align}




\begin{figure}
\includegraphics[width=0.5\textwidth]{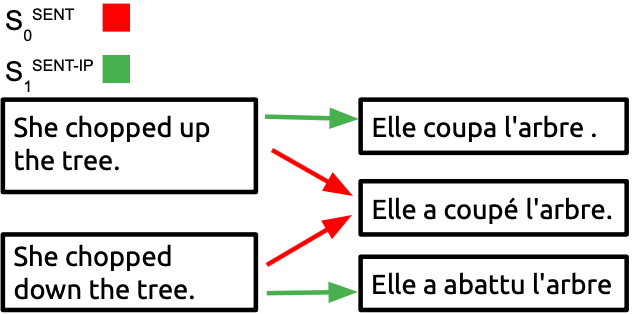}
\caption{State-of-the-art neural image captioner $\SZEROSENT$ loses a meaning distinction which \emph{informative} model $\SONESENTIP$ preserves.}
\label{figure:2}
\end{figure}

%
\paragraph{Survey} To evaluate the nature of this problem, we explored a corpus\footnote{Generated by selecting short sentences from the Brown corpus \citep{kuvcera1967computational}, translating them to German, and taking the best two candidate translations back into English, if these two themselves translate to a single German sentence. Translation in both directions was done with Fairseq.} of 500 pairs of distinct English sentences which map to a single German one (the  evaluation language in section \ref{eval}). We identify a number of common causes for the many-to-one maps. Two frequent types of verbal distinction lost when translating to German are tense (54 pairs, e.g. ``...others \{were, have been\} introduced .'') and modality (16 pairs, e.g. ``...prospects for this year \{\emph{could}, \emph{might}\} be better.''), where German ``k{\"o}nnen'' can express both epistemic and ability modality, distinguished in English with ``might'' and ``could'' respectively. Owing to English's large vocabulary, lexical difference in verb (31 pairs, e.g. ``arise'' vs. ``emerge'' ), noun (56 pairs, e.g. ``mystery'' vs. ``secret''), adjective (47 pairs, e.g. ``unaffected'' vs. ``untouched'') or deictic/pronoun (32 pairs, usually ``this'' vs ``that'') are also common.
A large number of the pairs differ instead either orthographically, or in other ways that do not correspond to a clear semantic distinction (e.g.  ``She had \{\emph{taken}, \emph{made}\} a decision.''). 





%



\paragraph{Our approach}
While languages differ in what distinctions they are \emph{required} to express, all are usually capable of expressing any given distinction when desired. As such, meaning loss of the kind discussed above is, in theory, avoidable.
To this end, we propose a method to reduce meaning loss by applying the Rational Speech Acts (RSA) model of an informative speaker to translation. RSA has been used to model natural language \emph{pragmatics} \citep{goodman2016pragmatic}, and recent work has shown its applicability to image captioning \citep{D16-1125,vedantam2017context,mao2016generation}, another sequence-generation NLP task.
Here we use RSA to define a translator which reduces many-to-one mappings and consequently meaning loss, in terms of a pretrained neural translation model.
We introduce a strategy for performing inference efficiently with this model in the setting of translation, and show gains in \emph{cycle-consistency}\footnote{Formally, say that a pair of functions $f : A\to B$, $g : B\to A$ is cycle-consistent if $g \cdot f = id$, the identity function. If $f$ is \emph{not} one-to-one, then $(f,g)$ is not cycle-consistent. (Note however that when $A$ and $B$ are infinite, the converse does not hold: even if $f$ and $g$ are both one-to-one, $(f,g)$ need not be cycle-consistent, since many-to-one maps between infinite sets are not necessarily bijective.)} as a result. 
Moreover, we obtain improvements in translation quality (BLEU score), demonstrating that the goal of meaning preservation directly yields improved translations.

\begin{figure}
\begin{tabular}{ll|ll}
 & $A$  & He is wearing glasses. &  \\
 & $B$ & He wears glasses. &  \\
  \hline
 & $\SZEROSENT(A)$ & Er trägt eine Brille. &  \\
 & $\SZEROSENT(B)$ & Er trägt eine Brille . &  \\
  \hline
 & $\SONESENTIP(A)$ & Er trägt jetzt eine Brille. &  \\
 & $\SONESENTIP(B)$ & Er hat eine Brille. & 
\end{tabular}
\caption{Similar to Figure \ref{figure:2},  $\SZEROSENT$ collapses two English sentences into a single German one, whereas $\SONESENTIP$ distinguishes the two in German.}
\label{figure:1}
\end{figure}

\section{Meaning Preservation as Informativity} 

In the RSA framework, speakers and listeners, modeled as Bayesian agents, reason about each other in a nested fashion. We refer to listeners and speakers which do not reason about another agent as $L_0$ and $S_0$ respectively, and an agent which reasons about another agent as $L_1$ or $S_1$. For instance, an informative speaker model $S_1$ is given a state $w \in W$, and chooses an utterance $u\in U$ to convey $w$ to $S_1$'s model of a listener. By contrast, $S_0$ chooses utterances without a listener model in mind --- its behavior might be determined by a semantics, or in our case, by a pretrained neural model.


For translation, the state space $W$ is a set of source language sentences (sequences of words in the language), while $U$ is a set of target language sentences. $S_1$'s goal is to choose a translation $u$ which allows a listener to pick out the source sentence $w$ from among the set of distractors. This informative behavior discourages many-to-one maps that a non-informative translation model $S_0$ might allow. 


\paragraph{$\mathbf{S_0}$ Model} BiLSTMs with attention \citep{bahdanau2014neural}, and more recently CNNs \citep{gehring2016convolutional} and entirely attention based models \citep{vaswani2017attention} constitute the state-of-the-art architectures in neural machine translation . All of these systems, once trained end-to-end on aligned data, can be viewed as a conditional distribution\footnote{We use $S_{0/1}^{\mathit{WD}}$ and $S_{0/1}^{\mathit{SNT}}$ respectively to distinguish word and sentence level speaker models } $\SZEROWORD(\mathit{wd}|w,\mathit{c})$, for a word \emph{wd} in the target language, a source language sentence $w$, and a partial sentence  $c$ in the target language. $\SZEROWORD$ yields a distribution $\SZEROSENT$ over full sentences\footnote{Python list indexing conventions are used, ``+'' means concatenation of list to element or list}: 
\begin{equation}
\SZEROSENT(u|w,\mathit{c}) = \prod_t \SZEROWORD(u[t]|w,\mathit{c}+u[:t]) \label{szerosent}
\end{equation}
$\SZEROSENT$ returns a distribution over continuations of $c$ into full target language sentences\footnote{In what follows, we omit \emph{c} when it is empty, so that $\SZEROSENT(u|w)$ is the probability of sentence $u$ given $w$}. To obtain a sentence from $\SZEROSENT$ given a source language sentence $s$, one can greedily choose the highest probability word from $\SZEROWORD$ at each timestep, or explore a beam of possible candidates. We implement $\SZEROWORD$ (in terms of which all our other models are defined) using \emph{Fairseq}'s publicly available\footnote{\url{https://github.com/pytorch/fairseq}} pretrained Transformer models for English-German, and for German-English train a CNN using \emph{Fairseq}.


\subsection{Explicit Distractors} We first describe a sentence level, \emph{globally pragmatic} model $\SONESENTGP$ for the simple case where a source language sentence needs to be distinguished from a presupplied distractor\footnote{Implementations for all models are available to \url{https://github.com/reubenharry/pragmatic-translation}} (as in the pairs shown in figures (\ref{figure:1}) and (\ref{figure:2})). We use this model as a stepping stone to one which requires an input sentence in the source language only, and no distractors. We begin by defining a listener $\LONESENT$, which receives a target language sentence $u$ and infers which sentence $w \in W$ (a presupplied set such as the pair (\ref{sent:3}) and (\ref{sent:4})) would have resulted in the pretrained neural model $\SZEROSENT$ producing $u$:

\begin{equation}
\LONESENT(w|u) \propto \frac{\SZEROSENT(u|w)}{\sum_{w'\in W}\SZEROSENT(u|w')} \label{l1sent}
\end{equation}

 This allows $\SONESENTGP$ to be defined in terms of $\LONESENT$, where $U$ is the set of all possible target language sentences\footnote{$\alpha$ is a hyperparameter of $\SONESENTGP$; as it increases, the model cares more about being informative and less about producing a reasonable translation.}:

\begin{equation}
\SONESENTGP(u|w) = \frac{\SZEROSENT(u|w)\LONESENT(w|u)^{\alpha}}{\sum_{u'\in U}\SZEROSENT(u'|w)\LONESENT(w|u')^{\alpha}} \label{s1sentgp}
\end{equation}

The key property of this model is that, for $W=\{A,B\}$, when translating $A$, $\SONESENTGP$ prefers translations of $A$ that are unlikely to be good translations of $B$. So for pairs like (\ref{sent:3}) and (\ref{sent:4}), $\SONESENTGP$ is compelled to produce a translation for the former that reflects its difference from the latter, and vice versa. 

\paragraph{Inference} Since $U$ is an infinite set, exactly computing the most probable utterance under $\SONESENTGP(\cdot | w)$ is intractable. \citet{D16-1125} and \citet{mao2016generation} perform approximate inference by sampling the subset of $U$ produced by a beam search from $\SZEROSENT$. \citet{vedantam2017context} and \citet{cohn2018pragmatically} employ a different method, using an \emph{incremental} model $\SONESENTIP$ as an approximation of $\SONESENTGP$ on which inference can be tractably performed.

$\SONESENTIP$ considers informativity not over the whole set of utterances, but instead at each decision of the next word in the target language sentence. For this reason, the incremental method avoids the problem of lack of beam diversity encountered when subsampling from $\SZEROSENT$, which becomes especially bad when producing long sequences, as is often the case in translation. $\SONESENTIP$ is defined as the product of informative decisions, specified by $\SONEWORD$ (itself defined in terms of $\LONEWORD$), which are defined analogously to (\ref{s1sentgp}) and (\ref{l1sent}).
    
%
\begin{equation}
\LONEWORD(w|\emph{wd},\emph{c}) \propto \SZEROWORD(\emph{wd}|w,\emph{c})
\end{equation}
\begin{align}
\SONEWORD(\mathit{wd}|w,\emph{c}) \propto  \\
        &\quad\SZEROWORD(\emph{wd}|w,\emph{c}) * 
            \LONEWORD(w|\emph{wd},\emph{c})^{\alpha} \nonumber
\end{align}
        
\begin{equation}
\SONESENTIP(u|w,\mathit{c}) = \prod_t \SONEWORD(u[t]|w,\mathit{c}+u[:t])
\end{equation}




\paragraph{Examples}
$\SONESENTIP$ is able to avoid many-to-one mappings by choosing more informative translations. For instance, its translation of (\ref{sent:3}) is ``Ces animaux courent vite'' (\emph{These} animals run fast.). 
See figures (\ref{figure:2}) and (\ref{figure:1}) for other examples of many-to-one mappings under $\SZEROSENT$ avoided by $\SONESENTIP$.



\subsection{Avoiding Explicit Distractors}

While $\SONESENTIP$ can disambiguate between pairs of sentences, it has two shortcomings. First, it requires one (or more) distractors to be provided, so translation is no longer fully automatic. Second, because the distractor set $W$ consists of only a pair (or finite set) of sentences, $\SONESENTIP$ only cares about being informative up to the goal of distinguishing between these sentences. Intuitively, total meaning preservation is achieved by a translation which distinguishes the source sentence $w$ from every other sentence in the source language which differs in meaning.
    
    
Both of these problems can be addressed by introducing a new ``cyclic'' globally pragmatic model $\SONESENTCGP$ which reasons not about $\LONESENT$ but about a pretrained translation model from target language to source language, which we term $\LZEROSENT$.
\begin{equation}
    \SONESENTCGP(u|w) \propto \SZEROSENT(u|w)\LZEROSENT(w|u)^{\alpha} 
\end{equation}

$\SONESENTCGP$ is like $\SONESENTGP$, but its goal is to produce a translation which allows a listener model (now $\LZEROSENT$) to infer the original sentence, not among a small set of presupplied possibilities, but among \emph{all source language sentences}. As such, an optimal translation $u$ of $w$ under $\SONESENTCGP$ has high probability of being generated by $\SZEROSENT$ and high probability of being back-translated to $w$ by $\LZEROSENT$. $\SONESENTCGP$ is very closely related to reconstruction methods, e.g. \citep{tu2017neural}.







\paragraph{Incremental Model} Exact inference is again intractable, though as with  $\SONESENTGP$, it is possible to approximate by subsampling from $\SZEROSENT$. This is very close to the approach taken by \citep{li2016simple}, who find that reranking a set of outputs by probability of recovering input ``dramatically decreases the rate of dull and generic responses.'' in a question-answering task. However, because the subsample is small relative to $U$, they use this method in conjunction with a diversity increasing decoding algorithm.
            

As in the case with explicit distractors, we instead opt for an incremental model, now $\SONESENTCIP$ which approximates $\SONESENTCGP$. The definition of $\SONESENTCIP$ (\ref{sentcgp}) is more complex than the incremental model with explicit distractors ($\SONESENTIP$) since $\LZEROWORD$ must receive complete sentences, rather than partial ones like $\LONEWORD$. As such, we need to marginalize over continuations $k$ of partial sentences in the target language:
\begin{align}
&\SONEWORDC(\mathit{wd}|w,c) \nonumber \propto \SZEROWORD(\mathit{wd}|w,c)*\\&\sum_{k}( \LZEROSENT(w|c+\mathit{wd}+k) \SZEROSENT(k|w,c+\mathit{wd})) \label{marg}
\end{align}

\begin{equation}
\SONESENTCIP(u|w,\mathit{c})=\prod_t \SONEWORDC(u[t]|w,\mathit{c}+u[:t]) \label{sentcgp}
\end{equation}

Since the sum over continuations of $c$ in (\ref{marg}) is intractable to compute exactly, we approximate it by a single continuation, obtained by greedily unrolling $\SZEROSENT$. The whole process of generating a new word $\mathit{wd}$ of the translation from a sequence $c$ and a source language sentence $w$ is as follows: first use $\SZEROWORD$ to generate a set of candidates for the next word (in practice, we only consider two, for efficiency). For each of these, use $\SZEROSENT$ to greedily unroll a full target language sentence from $c+\mathit{wd}$, namely $c+\mathit{wd}+k$, and rank each $\mathit{wd}$ by the probability $\LZEROSENT(w|c+\mathit{wd}+k)$.




\subsection{Evaluating the Informative Translator} \label{eval}

Our objective is to improve meaning preservation without detracting from translation quality in other regards (e.g. grammaticality). We conduct our evaluations on English to German translation, making use of publicly available pre-trained English-German and German-English Fairseq models. The pragmatic model we evaluate is $\SONESENTCIP$ since, unlike $\SONESENTIP$, it is not necessary to hand-supply a distractor set of source language sentences.

An example of the behavior of $\SONESENTCIP$ and $\SZEROSENT$ on of our test sentences is shown below; $\SZEROSENT$ is able to preserve the phrase ``world's eyes'', which $\SZEROSENT$ translates merely as ``world'':

\begin{itemize}
\item Source sentence: Isolation keeps the world's eyes off Papua.
\item Reference translation: Isolation h\"alt die Augen der Welt fern von Papua.
\item $\SZEROSENT$: Die Isolation h\"alt die Welt von Papua fern.
\item $\SONESENTCIP$: Die Isolation h\"alt die Augen der Welt von Papua fern.
\end{itemize}

We use cycle-consistency as a measure of meaning preservation, since the ability to recover the original sentence requires meaning distinctions not to be collapsed. In evaluating cycle-consistency, it is important to use a separate target-source translation mechanism than the one used to define $\SONESENTCIP$. Otherwise, the system has access to the model which evaluates it and may improve cycle-consistency without producing meaningful target language sentences. For this reason, we translate German sentences (produced by $\SZEROSENT$ or $\SONESENTCIP$) back to English with \emph{Google Translate}. To measure cycle-consistency, we use the BLEU metric (implemented with sacreBLEU \citep{post2018call}), with the original sentence as the reference.

However, this improvement of cycle consistency, especially with a high value of $\alpha$, may come at the cost of translation quality. Moreover, it is unclear whether BLEU serves as a good metric for evaluating sentences of a single language. To further ensure that translation quality is not compromised by $\SONESENTCIP$, we evaluate BLEU scores of the German sentences it produces. This requires evaluation on a corpus of aligned sentences, unlike the sentences collected from the Brown corpus in section \ref{brown}\footnote{While we find that $\SONESENTCIP$ improves cycle-consistency for the Brown corpus over $\SZEROSENT$, we have no way to establish whether this comes at the cost of translation quality.}.

We perform both evaluations (cycle-consistency and translation) on 750 sentences\footnote{Our implementation of $\SONESENTCIP$ was not efficient, and we could not evaluate on more sentences for reasons of time.} of the 2018 English-German WMT News test-set.\footnote{\url{http://www.statmt.org/wmt18/translation-task.html}} We use greedy unrolling in all models (using beam search is a goal for future work). For $\alpha$ (which represents the trade-off between informativity and translation quality) we use $0.1$, obtained by tuning on validation data.

\paragraph{Results} As shown in table (\ref{table:1}), $\SONESENTCIP$ improves over $\SZEROSENT$ not only in cycle-consistency, but in translation quality as well. This suggests that the goal of preserving information, in the sense defined by $\SONESENTCGP$ and approximated by $\SONESENTCIP$, is important for translation quality.



\begin{table}[t!]
\begin{center}
\setlength{\tabcolsep}{12pt}
\begin{tabular}{l l l l}
\toprule 
\bf Model & \bf Cycle & \bf Translate \\ 
\midrule
$\SZEROSENT$  & 43.35 & 37.42 \\
$\SONESENTCIP$ & \textbf{47.34} & \textbf{38.29} \\
\bottomrule
\end{tabular}
\end{center}
\caption{BLEU score on cycle-consistency and translation for WMT, across baseline and informative models.  Greedy unrolling and $\alpha=0.1$}

\label{table:1}
\end{table}

\section{Conclusions} 
We identify a shortcoming of state-of-the-art translation systems and show that a version of the RSA framework's informative speaker $S_1$, adapted to the domain of translation, alleviates this problem in a way which improves not only cycle-consistency but translation quality as well.
The success of $\SONESENTCIP$ on two fairly similar languages raises the question of whether improvements will increase for more distant language pairs, in which larger scale differences exist in what information is obligatorily represented - this is a direction for future work.

\section{Acknowledgements}

Thanks to the reviewers for their substantive comments, and to Daniel Fried and Jacob Andreas for many helpful discussions during the development of this project.


   

\bibliography{naaclhlt2019}

\begin{thebibliography}{12}
\expandafter\ifx\csname natexlab\endcsname\relax\def\natexlab#1{#1}\fi

\bibitem[{Andreas and Klein(2016)}]{D16-1125}
Jacob Andreas and Dan Klein. 2016.
\newblock \href {http://aclweb.org/anthology/D16-1125} {Reasoning about
  pragmatics with neural listeners and speakers}.
\newblock In \emph{Proceedings of the 2016 Conference on Empirical Methods in
  Natural Language Processing}, pages 1173--1182. Association for Computational
  Linguistics.

\bibitem[{Bahdanau et~al.(2014)Bahdanau, Cho, and Bengio}]{bahdanau2014neural}
Dzmitry Bahdanau, Kyunghyun Cho, and Yoshua Bengio. 2014.
\newblock Neural machine translation by jointly learning to align and
  translate.
\newblock \emph{arXiv preprint arXiv:1409.0473}.

\bibitem[{Cohn-Gordon et~al.(2018)Cohn-Gordon, Goodman, and
  Potts}]{cohn2018pragmatically}
Reuben Cohn-Gordon, Noah Goodman, and Christopher Potts. 2018.
\newblock \href {http://aclweb.org/anthology/N18-2070} {Pragmatically
  informative image captioning with character-level inference}.
\newblock In \emph{Proceedings of the 2018 Conference of the North American
  Chapter of the Association for Computational Linguistics: Human Language
  Technologies, Volume 2 (Short Papers)}, pages 439--443. Association for
  Computational Linguistics.

\bibitem[{Gehring et~al.(2016)Gehring, Auli, Grangier, and
  Dauphin}]{gehring2016convolutional}
Jonas Gehring, Michael Auli, David Grangier, and Yann~N Dauphin. 2016.
\newblock A convolutional encoder model for neural machine translation.
\newblock \emph{arXiv preprint arXiv:1611.02344}.

\bibitem[{Goodman and Frank(2016)}]{goodman2016pragmatic}
Noah~D Goodman and Michael~C Frank. 2016.
\newblock Pragmatic language interpretation as probabilistic inference.
\newblock \emph{Trends in Cognitive Sciences}, 20(11):818--829.

\bibitem[{Ku{\v{c}}era and Francis(1967)}]{kuvcera1967computational}
Henry Ku{\v{c}}era and Winthrop~Nelson Francis. 1967.
\newblock \emph{Computational analysis of present-day American English}.
\newblock Dartmouth Publishing Group.

\bibitem[{Li et~al.(2016)Li, Monroe, and Jurafsky}]{li2016simple}
Jiwei Li, Will Monroe, and Dan Jurafsky. 2016.
\newblock A simple, fast diverse decoding algorithm for neural generation.
\newblock \emph{arXiv preprint arXiv:1611.08562}.

\bibitem[{Mao et~al.(2016)Mao, Huang, Toshev, Camburu, Yuille, and
  Murphy}]{mao2016generation}
Junhua Mao, Jonathan Huang, Alexander Toshev, Oana Camburu, Alan~L Yuille, and
  Kevin Murphy. 2016.
\newblock Generation and comprehension of unambiguous object descriptions.
\newblock In \emph{Proceedings of the IEEE conference on computer vision and
  pattern recognition}, pages 11--20.

\bibitem[{Post(2018)}]{post2018call}
Matt Post. 2018.
\newblock A call for clarity in reporting bleu scores.
\newblock \emph{arXiv preprint arXiv:1804.08771}.

\bibitem[{Tu et~al.(2017)Tu, Liu, Shang, Liu, and Li}]{tu2017neural}
Zhaopeng Tu, Yang Liu, Lifeng Shang, Xiaohua Liu, and Hang Li. 2017.
\newblock Neural machine translation with reconstruction.
\newblock In \emph{Thirty-First AAAI Conference on Artificial Intelligence}.

\bibitem[{Vaswani et~al.(2017)Vaswani, Shazeer, Parmar, Uszkoreit, Jones,
  Gomez, Kaiser, and Polosukhin}]{vaswani2017attention}
Ashish Vaswani, Noam Shazeer, Niki Parmar, Jakob Uszkoreit, Llion Jones,
  Aidan~N Gomez, {\L}ukasz Kaiser, and Illia Polosukhin. 2017.
\newblock Attention is all you need.
\newblock In \emph{Advances in Neural Information Processing Systems}, pages
  5998--6008.

\bibitem[{Vedantam et~al.(2017)Vedantam, Bengio, Murphy, Parikh, and
  Chechik}]{vedantam2017context}
Ramakrishna Vedantam, Samy Bengio, Kevin Murphy, Devi Parikh, and Gal Chechik.
  2017.
\newblock Context-aware captions from context-agnostic supervision.
\newblock In \emph{Computer Vision and Pattern Recognition (CVPR)}, volume~3.

\end{thebibliography}
\bibliographystyle{acl_natbib}
\end{document}